# Missing Data using Decision Forest and Computational Intelligence

D. Moon and T. Marwala
School of Electrical and Information Engineering, University of Witwatersrand, Private Bag x5, Wits, 2050
South Africa

*Abstract—* **Autoencoder Neural Network is implemented to estimate the missing data. Genetic Algorithm (GA) is implemented for network optimization and estimating the missing data. Missing data is treated as Missing At Random (MAR) by implementing maximum likelihood algorithm. The network performance is determined by calculating the network's Mean Square Error (MSE). The network is further optimized by implementing Decision Forest (DF). The impact of missing data is then investigated on both ANN-GA and ANN-GA-DF network.**

## I. INTRODUCTION

Missing data poses problems when performing data analysis. Complete and accurate data are necessary in obtaining good inference [1]. There are applications which require missing data to be estimated [1]. When estimating the data, data needs to be as close to, if not the same, the original value. In Autoencoder Neural Network (ANN), change in one input affects every outcome as illustrated in figure 1. Therefore, it is vital to estimate these missing data as accurately as possible. Different types of missing data are explained in the paper. There are different ways of dealing with each missing data type. Some methods are more effective than others in dealing with certain types of missing data. In this paper, missing data from given database is assumed to be Missing At Random (MAR) and is estimated using maximum likelihood algorithm. ANN in conjunction with GA is implemented to construct neural network to estimate missing data. Maximum likelihood is implemented at GA by replacing the genetic fitness function. The network is further optimized by the addition of DF. Accuracy of the estimation and the overall performance were assessed by calculating MSE at each stage of the constructions. Mean value of each variables were also used to assess the performance as well as the impact of missing data. Findings also include impact on the network performance by implementation of DF. There are no specific performance criteria to be achieved except for a successful system impact analysis. The database provided is from *National HIV and Syphilis Sero-Prevalence survey* [2].

## II. BACKGROUND

### A. Types of missing data

There are mainly 3 types of missing data: Missing At Random (MAR), Missing Completely At Random (MCAR) and Missing Not At Random (MNAR) [3]. MAR is when the missing data is dependant on other variable in the dataset [3]. In other words, an incorrect data for a variable can be the cause of another variable's data to be missing. The missing data pattern is traceable in MAR [3]. MCAR is when the missing data has no dependence on other variable/s or even to itself [3]. Data is simply "just missing" and no relation can be derived between variables to determine the cause of the missing data. MNAR is when missing data is dependant on other missing data in the dataset and is un-ignorable [2]. Definitions of these terms are also found in [1], [4], [5], [6]. The missing data in the dataset used are treated as MAR by implementing maximum likelihood algorithm. This is explained further in later sections.

### B. Autoencoder Neural Network (ANN)

Autoencoder Neural Network is Autoassociative Neural Network encoder [7]. In Autoassociative Neural Network, number of hidden nodes is less than the number of input nodes. ANN predicts input as an output. Therefore, ANN is trained to recall the inputs as outputs [7]. Fewer hidden nodes compared to input nodes characterizes bottleneck like feature as seen in figure 1. This allows redundant data to be removed because the network inputs will be projected to a smaller space [7]. Hidden node is single layer because it is adequate for MLP according to universal approximation theory [7].

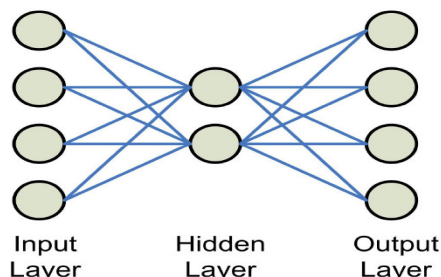

Figure 1: Schematic diagram of ANN architecture [2]

ANN architecture is identical to that of Multi-Layered Perceptron (MLP), except for the difference in number of hidden nodes in ANN. While MLP can have more number of hidden nodes than its input nodes, ANN has less number of hidden nodes than input nodes. Since MLP is used to construct ANN, same types of training functions can be used in ANN as would in MLP.

*C. Genetic Algorithm (GA)*

Genetic Algorithm is optimization theorem based on biological evolution [8]. Genetic Algorithm computes a fitness of all the values randomly created in a population. GA uses the value with highest fitness to derive new values. Fitness function needs to be derived before fitness of values can be determined. Fitness function will be the maximum likelihood algorithm. See next section for maximum likelihood description. GA determines the fitness of each value in the generated population by assessing the value using the fitness function. GA then performs mutation and crossover functions for specified number of generations on the fittest value chosen. The best solution (value with the highest fitness value) is produced in the end.

The best solution is the estimated data and replaces the missing data. In general, the fittest individuals of any population tend to reproduce and survive to the next generation, thus improving with successive generations. However, inferior individuals can, by chance, survive and also reproduce [9]. Genetic algorithms have been shown to solve linear and non-linear problems by exploring all regions of the state space and exponentially exploiting promising areas through mutation, crossover, and selection operations applied to individuals in the population [9]. GA has shown to be affective in estimating missing data in [1], [3], [4], [6], [7].

*D. Algorithm for estimating missing data*

There are numerous ways for estimating or treating missing data in database. There are mainly two approaches to missing data. Delete or estimate the missing data. Deleting missing data is plausible if there are few records with missing data. If there are many missing data then estimating the missing data is the more plausible approach because the ANN requires certain amount of data to train.

Some of the missing data estimation methods include mean substitution, regression methods, hot deck imputation, Expectation Maximization, maximum likelihood and multiple imputation [1]. Two most widely used methodologies are EM and maximum likelihood [3], [4], [6], [7].

Maximum likelihood is good application in imputing the missing values because it is based on precise statistical model of the data [3]. Maximum likelihood uses constructs best 1st and 2nd order moment estimates under the MAR assumption [1]. Since the missing data is treated as MAR, maximum likelihood is used.

In maximum likelihood, there are two variables: $\vec{X}_k$ =known data and $\vec{X}_u$ =unknown data. The error is calculated as in equation 1.

$$error = \begin{bmatrix} \vec{X_K} \\ \vec{X_U} \end{bmatrix} - F\left\{ \begin{bmatrix} \vec{X_K} \\ \vec{X_U} \end{bmatrix} \right\} \quad (1)$$

The matrix in equation 1 is a single record of a data. Second part of the equation is the output of a network with $\vec{X}_k$ and $\vec{X}_u$ passed through it. The error function is squared to avoid negative error as is shown in equation 2. The GA will estimate the $\vec{X}_u$, and pass it through ANN. The GA chooses the highest value of its evaluation function in choosing the "fittest value" [9]. Desired value is the one with the lowest error so error determined by equation 2 is converted to negative value and used as an evaluation function in GA.

$$error = \left( \begin{bmatrix} \vec{X_K} \\ \vec{X_U} \end{bmatrix} - F\left\{ \begin{bmatrix} \vec{X_K} \\ \vec{X_U} \end{bmatrix} \right\} \right)^2 \quad (2)$$

As seen in figure 1, every input variable are passed through each hidden nodes. Therefore, change in one input affects all outputs. Minimizing the error in equation 2 ensures that outputs are as close to the inputs and that output variables of ANN are not affected. This also ensures that the estimated data are close to the actual data. This method of ANN with GA optimization is preferred to the use of classifiers because feedback mechanism is applied [7]. The benefit of feedback system is that it is less subjective to the noise in dataset [7]. The overall system algorithm is illustrated in figure 2. Refer to [8] for illustration of using classifier to estimate data.

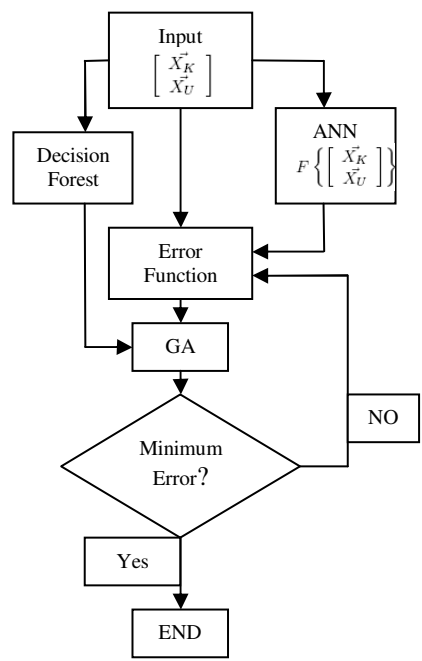

Figure 2: Missing data estimation algorithm

*E. Decision Forest*

To understand decision forest, we will first look at single decision tree. Decision tree is a classifier in a tree form [4]. It shows all possible outcomes by using 'if...then' format [11]. Decision tree has leaves and decision nodes [11]. Each leaf indicates a class and each decision node specifies some test to be carried out on a single attribute value, with one branch and sub tree for each possible outcome of the test [11]. Decision tree method as explained in [11] is provided below.

- Begin at the root of the tree and move through it until a leaf is encountered.
- At each non-leaf node, case's outcome for the test at the node is determined and attention shifts to the root of the sub tree corresponding to this outcome.
- When a leaf is reached, the class of the case is predicted to be that recorded at the leaf.

There are number of algorithms to construct a decision tree. These include ID3 and C4.5 [11], classification and regression tree [12] and OC1 [13]. Classification and regression algorithm is used in this investigation. In classification and regression, a leaf indicates a number and not a class. Just as the name suggests, decision forest is built by large number of decision trees. The concept behind decision forest is that more number of trees gives more accurate estimate than single tree by providing more nodes and leaves. Trees can be trained using different algorithms, predictors or subsets. Tree boosting also uses number of trees. Tree boosting is not quite the same as decision forest but can be considered as decision forest in a sense that it has more than one tree. In tree boosting each tree is constructed to reduce the error from the previous tree to produce less error value [14].

### III. PREPROCESSING DATA

The ANN requires fully recorded dataset to train. The dataset must not have missing data, outliers or incorrect data. The missing data and outliers were treated using listwise/casewise data deletion for training set. The listwise/casewise data deletion is explained in [1]. If the recorded data are incorrect but is within the correct bounds, it can not be identified. For example; if a participant enters an incorrect age, 32 instead of 29, ANN can not detect that it is incorrect because it is still within the correct range. It is assumed that unknown percentage of data used could be incorrect data. However, normalizing the entire dataset to be between 0 and 1 reduces the probability of these data marginalizing the network. The variables and their correct range are listed in table 1.

The variables (province, region and race) with text data had to be converted to integers for Matlab to use them. The ranges of father's and mother's age were assumed as in table 1. Number of successful pregnancy can not exceed number of pregnancy so data with such values were considered incorrect data and was removed.

Table 1
Dataset variables and corresponding bounds

| Variable | Range/ Data type | Explanation |
|---|---|---|
| Province | Integers | Province of residence |
| Region | Integers | Region of residence |
| Age | 15-50 | Age of mother |
| Race | Integers | Race of mother |
| Education | 0-13 | Education level of mother |
| Gravidity | 0-8 | Number of pregnancy |
| Parity | 0-7 | Number of successful pregnancy |
| Father's age | 16-65 | Age of baby's father |
| HIV status | 0 or 1 | HIV status of a mother |
| RPR | 0 or 1 | Rapid Plasma Reagin |

"*Rapid Plasma Reagin (RPR) is a screening test for syphilis; it looks for antibodies that are present in the blood of people who have the disease*" [15]. Therefore, 1 means antibodies present and 0 means antibodies not present. In HIV status, 0 means negative and 1 means positive. After the removal of missing data, outliers and some incorrect data, the dataset was reduced from 16738 to 11980. The final dataset was broken down into training, validation and testing dataset for ANN. Normalization was implemented by using equation 1.

$$\bar{x} = \frac{x - x_{min}}{x_{max} - x_{min}} \quad (1)$$

Where, $\bar{x}$ is normalized value.

### IV. NETWORK PARAMETER

*A. Autoencoder Neural Network (ANN)*

ANN was modeled using Multi-Layered Perceptron. The optimal parameters were chosen to produce the lowest MSE. The chosen parameters are listed in table 2.

Table 2
Auto-associative neural network parameters

| Input nodes | 10 |
|---|---|
| Output nodes | 10 |
| Hidden nodes | 9 |
| Output function | Linear |
| Training function | Short Conjugate Gradient |
| Training cycle | 1000 |
| Error | $6.3747 \times 10^{-11}$ |

Number of training cycle was chosen by using the early stopping method. Each time output or training function was changed the mean square error was recalculated to determine the optimal function. The MSE was calculated using the validation dataset.

### B. Genetic Algorithm (GA)

Genetic algorithm was implemented to optimize the estimation of missing data as correctly as possible. Ideally, data estimated by GA should be the same as the target value and not affect other variables. Different combinations of parameters were tried and the parameter with lowest MSE value produced was chosen. Unfortunately, the final parameters included large number of population and generation. This compromised the investigation with extreme computation time. Initial computation of data estimation was unsuccessful due to the fact that the computation took over a day and had to be disrupted without producing a result. Despite realizing that MSE decreased with increase in number of generations, accuracy had to be compromised for efficiency in computation period. Since thousands of data had to be dealt with, number of generation was reduced to 10 for more realistic computation time. This was done in the assumption that bound provided by decision forest later on will significantly increase the accuracy without compromising the computation time. The parameters chosen are listed in table 3.

Table 3
Genetic Algorithm parameters

| Initial population size | 15 |
|---|---|
| Number of generation | 10 |
| Crossover function | Simple crossover |
| Mutation function | Boundary mutation |

### C. Decision Forest (DF)

Decision trees used to induce decision forest are regression model. Trees were constructed using Matlab built-in functions. The forest contained four trees. This is not really a forest as forest ideally contains hundreds of trees. This model, however, does implicate the concept of forest. The values at each relevant leaf from all trees were compared to derive the minimum and a maximum bound to be passed into the GA. Intelligence behind this is that the bounds provided by decision forest will include the global minimum.

It was found that the final trees built with large number of data are very complicated. Trees can be pruned to produce simplified trees. This is done by discarding one or more sub-trees and replacing them with leaves [11]. Pruned trees produced higher error rate than a simple tree and was not implemented.

When dealing with large dataset and indirect method of growing trees are required, technique called windowing can be used. This was implemented to increase accuracy and save virtual memory of the construction instrument used.

Windowing simply is a technique to construct a tree to be able to classify cases that was not included in the training set [11]. Downside to windowing is that it slows down the tree building process [11]. Windowing was implemented in following manner:

- Randomly select a subset of the training cases called "window" and develop decision tree from it.
- Classify training cases not included in the window
- Add misclassified cases into the window and develop $2^{nd}$ tree from it and test on remaining sets.
- Continue the steps until a tree built can classify those outside the window correctly.

## V. TEST AND RESULTS

Two criteria have been used to measure the performance of implemented design. First, accuracy of estimating missing data with and without decision forest was determined. The estimated data was compared to the actual data by plotting the actual data and estimated data. MSE was also used for determining the accuracy. Second, the impact of estimating missing data on the rest of the system was assessed.

The missing variable used was *education*. This data was the most plausible as it had higher range to test the accuracy of the estimation compared to HIV which only have 0 or 1. If the system outputs 0 or 1 for 70% of time then the system has a high probability of estimating the missing data correctly 70% of the time without actually estimating. The normalized data were returned to its original range as given in table 1 before MSE and mean values were calculated.

As illustrated in figure 1, change in one variable affects rest of the variables. Therefore, the system with the lowest MSE for data estimation and impact of missing data is the favorable of the two.

Figures 3 and 4 shows the target value plotted against estimated data by ANN-GA and ANN-GA-DF respectively. It is very clear from figures 3 and 4 that the estimated values from ANN-GA-DF are more accurate and close to the target value than that of ANN-GA. This conclusion is further substantiated by figure 5 and the data from table 4.

Figure 5 shows the error calculated per sample between the target value and estimated value by both systems. Error plotted in figure 5 shows clearly that ANN-GA has much higher error than ANN-GA-DF per every sample.

Although the mean of estimated data, recorded in table 4, from ANN-GA-DF system is further away from the target compared to that of ANN-GA, MSE is much lower for ANN-GA-DF. This validates the assumption made earlier that implementation of DF will further optimize the network's performance.

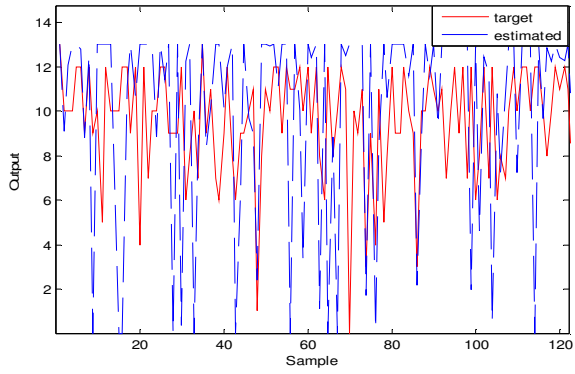

Figure 3: ANN-GA estimated vs. target data

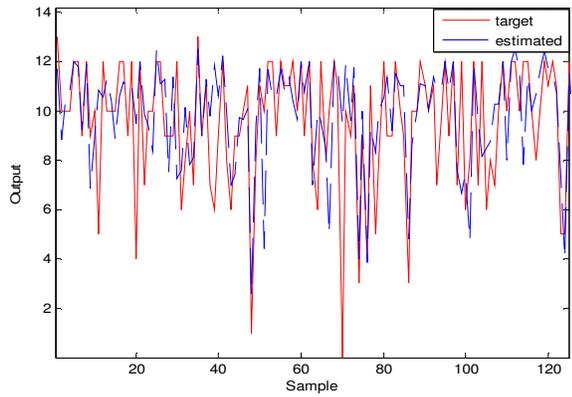

Figure 4: ANN-GA-DF estimated vs. target data

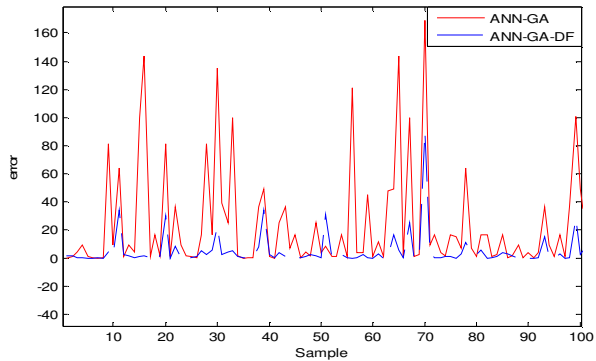

Figure 5: Error of 2 systems' estimated data

Table 4
Test results of two systems

|  | Target | ANN-GA | ANN-GA-DF |
|---|---|---|---|
| MSE | 0 | 22.4958 | 6.2746 |
| Mean | 9.4620 | 9.9989 | 10.4964 |

Dataset including the estimated education values were passed through ANN to assess the impact of estimated data on the rest of the dataset. The resulting dataset was compared to the input dataset. The aim of this approach is to examine the degree of change in other variables. Figure 6 shows the error per sample between input and output from ANN. Two input dataset are produced from ANN-GA and ANN-GA-DF. Error in figure 6 is average error of all variables.

Figure 6 and table 5 and 6 illustrates the minimal difference in error between the two systems. Both ANN-GA and ANN-GA-DF experiences almost identical impact from the missing data. Tests on both set of resulting data from ANN shows pretty identical MSE and each variable's mean value for both ANN-GA and ANN-GA-DF. Careful examination of values in table 5 and 6 reveal that ANN-GA-DF is less vulnerable to the impact of the missing data. However, the benefit is minimal. The impact of missing data in both systems is not significant because table 5 and 6 shows that both systems produce low MSE and the mean value of each variable is very close to the target.

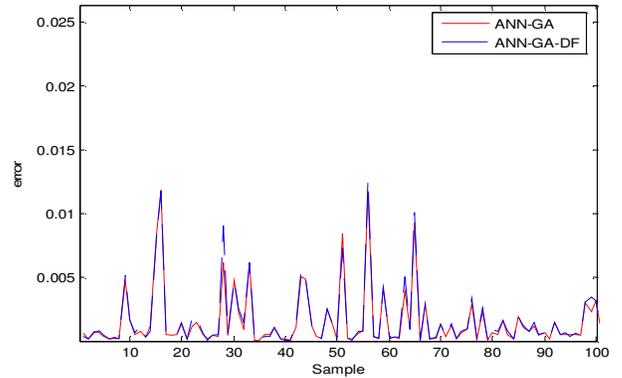

Figure 6: Error between input and output of 2 systems

Table 5
Missing data impact assessment result of ANN-GA

| Variables | MSE ($10^{-3}$) | Mean | Target Mean |
|---|---|---|---|
| Province | 3.7657 | 152.5 | 152.5 |
| Region | 2517.4 | 677.89 | 677.73 |
| Age | 4.2129 | 24.924 | 24.825 |
| Race | 2.8774 | 136.99 | 137 |
| Education | 10232 | 10.002 | 9.9989 |
| Gravidity | 7.0482 | 2.1127 | 2.127 |
| Parity | 3.9995 | 1.0019 | 0.992 |
| Father | 5.3255 | 29.724 | 29.725 |
| HIV | 0.00074002 | 0.22002 | 0.22 |
| RPR | 0.0037652 | 0.24907 | 0.025 |

Table 6
Missing data impact assessment result of ANN-GA-DF

| Variables | MSE ($10^{-3}$) | Mean | Target Mean |
|---|---|---|---|
| Province | 4.3501 | 152.49 | 152.5 |
| Region | 3.0706 | 677.86 | 677.73 |
| Age | 5.2538 | 24.92 | 24.825 |
| Race | 2.7853 | 137 | 137 |
| Education | 10820 | 10.502 | 10.4964 |
| Gravidity | 6.9267 | 2.1134 | 2.127 |
| Parity | 3.884 | 1.0039 | 0.992 |
| Father | 6.0166 | 29.73 | 29.725 |
| HIV | 0.00073213 | 0.22007 | 0.22 |
| RPR | 0.0039443 | 0.025261 | 0.025 |

Most significant feature observed by adding decision forest to the ANN-GA is the search bound. It is feasible that the same performance could be reached without adding decision forest by increasing population size and number of generations in genetic algorithm implementation. This increases the computation time dramatically. It would have taken ANN-GA over a day to complete the computation with the same performance as ANN-GA-DF. ANN-GA-DF increases the performance of network system without increasing the computation time.

Decision forest also helps in finding the global minimum by providing the range GA should search values for. For example; within the normalized range values (0-1), there could be number of minimal points (points where maximum likelihood error is at minimum). GA has no way of knowing which the global minimum is. Decision forest provides the range that the missing data is likely to be in. GA then search for the value that gives the minimal maximum-likelihood error within this range. The final data derived will be the global minimum.

## VI. CONCLUSION

Accuracy of estimation of missing data by Autoencoder Neural Network (ANN) was investigated. The ANN was optimized with Genetic Algorithm (GA). Maximum likelihood algorithm was implemented in GA for optimization. The performance of the designed network was measured in terms of Mean Square Error (MSE) and each variable's mean value was compared to the target mean values. The network was further optimized with implementation of Decision Forest (DF). MSE and mean values were calculated and compared to that of ANN-GA.

The results showed that ANN-GA-DF was better in estimating missing data as close to the original data than ANN-GA. This was possible as DF provided range of values for GA to find the global minimum. Impact of missing data on both systems was also investigated. It was found that both systems were impacted to the similar degree. This was due to good optimization of ANN which has good efficiency in predicting the input data as output.